\documentclass{article}

\usepackage[preprint]{iai_neurips_2024}

\usepackage[utf8]{inputenc} % allow utf-8 input
\usepackage[T1]{fontenc}    % use 8-bit T1 fonts
\usepackage{hyperref}       % hyperlinks
\usepackage{url}            % simple URL typesetting
\usepackage{booktabs}       % professional-quality tables
\usepackage{amsmath, amssymb}
\usepackage{amsfonts}       % blackboard math symbols
\usepackage{nicefrac}       % compact symbols for 1/2, etc.
\usepackage{microtype}      % microtypography
\usepackage{xcolor}         % colors
\usepackage{graphicx}
\usepackage{cleveref}

\setcitestyle{authoryear,open={(},close={)}}

\usepackage{theorem}

\newtheorem{definition}{Definition}[section]

\title{Interpretability as Compression: Reconsidering SAE Explanations of Neural Activations with MDL-SAEs}

\author{%
  Kola Ayonrinde\thanks{These authors contributed equally to this work.}\\
  MATS \\
  \texttt{koayon@gmail.com} \\
  \And
  Michael T. Pearce\footnotemark[1]\\
  MATS \\
  \texttt{michaelttpearce@gmail.com} \\
  \And
  Lee Sharkey \\
  Apollo Research \\
  \texttt{lee@apolloresearch.ai} \\
}

\begin{document}

\maketitle

\begin{abstract}
Sparse Autoencoders (SAEs) have emerged as a useful tool for interpreting the internal representations of neural networks. 
However, naively optimising SAEs for reconstruction loss and sparsity results in a preference for SAEs that are extremely wide and sparse. 
We present an information-theoretic framework for interpreting SAEs as lossy compression algorithms for communicating explanations of neural activations. We appeal to the Minimal Description Length (MDL) principle to motivate explanations of activations which are both accurate and concise. We further argue that interpretable SAEs require an additional property, “independent additivity”: features should be able to be understood separately.
We demonstrate an example of applying our MDL-inspired framework by training SAEs on MNIST handwritten digits and find that SAE features representing significant line segments are optimal, as opposed to SAEs with features for memorised digits from the dataset or small digit fragments.
We argue that using MDL rather than sparsity may avoid potential pitfalls with naively maximising sparsity such as undesirable feature splitting and that this framework naturally suggests new hierarchical SAE architectures which provide more concise explanations.
\end{abstract}

\section{Introduction}

Sparse Autoencoders (SAEs) \citep{le2013building, makhzani2013k} were developed to learn a dictionary of sparsely activating features describing a dataset. They have recently become popular tools for interpreting the internal activations of large foundation language models, often finding human-understandable features \citep{sharkey2022, huben2024sparse, bricken2023monosemanticity}.
% It has been speculated that the features that SAEs find may closely map onto the constituent features that the model is using.

Interpretability, in particular human-understandability, is difficult to optimise for since ratings---from humans or auto-interpretability methods \citep{bills2023language}---are not differentiable at training time and often cannot be efficiently obtained. Researchers often use sparsity, the number of nonzero feature activations as measured by the $L_0$ norm, as a proxy for interpretability. SAEs are typically trained with an additional $L_1$ penalty in their loss function to promote sparsity. 

We adopt an information theoretic view of SAEs, inspired by \cite{grunwald2007minimum}, which views SAEs as explanatory tools that compress neural activations into communicable explanations. This view suggests that sparsity may appear as a special case of a larger objective: minimising the description length of the explanations. This operationalises Occam's razor for selecting explanations: \textit{all else equal, prefer the more concise explanation}.

We introduce this information theoretic view by describing how SAEs can be used in a communication protocol to transmit neural activations. We then argue that interpretability requires explanations to have the property of independent additivity, which allows individual features to be interpreted separately and discuss SAE architectures that are compatible with this property. We find that sparsity (i.e. minimizing $L_0$) is a key component of minimizing description length but there are cases where sparsity and description length diverge. In these cases, minimizing description length directly gives more intuitive results. We demonstrate our approach empirically by finding the Minimal Description Length solution for SAEs trained on the MNIST dataset.

\section{SAEs are communicable explanations}
\label{sec:explanations}

SAEs aim to provide explanations of neural activations in terms of "features"\footnotemark{}. Here we reformulate SAEs as solving a communication problem: suppose that we would like to transmit the neural activations $x$ to a friend with some tolerance $\varepsilon$, either in terms of the reconstruction error or change in the downstream cross-entropy loss. Using the SAE as an encoding mechanism, we can approximate the representation of the activations in two parts. 
\textit{First}, we send them the SAE encodings of the activations $z = Enc(x)$. 
\textit{Second}, we send them a decoder network $Dec(\cdot)$ that recompiles these activations back to (some close approximation of) the neural activations, $\widehat{x} = Dec(z)$.

This is closely analogous to \textit{two-part coding schemes} \citep{grunwald2007minimum} for transmitting a program via its source code and a compiler program that converts the source code into an executable format. Together the SAE activations and the decoder provide an \textbf{explanation} of the neural activations, based on the definition below.  

\footnotetext{Here we use the term "feature" as is common in the literature to refer to a linear direction which corresponds to a member of the set of a (typically overcomplete) basis for the activation space. Ideally the features are relatively monosemantic and correspond to a single (causally relevant) concept. We make no guarantees that the features found by an SAE are the "true" generating factors of the system.}

\begin{figure}[h!]
\centering
\includegraphics[width=0.6\textwidth]{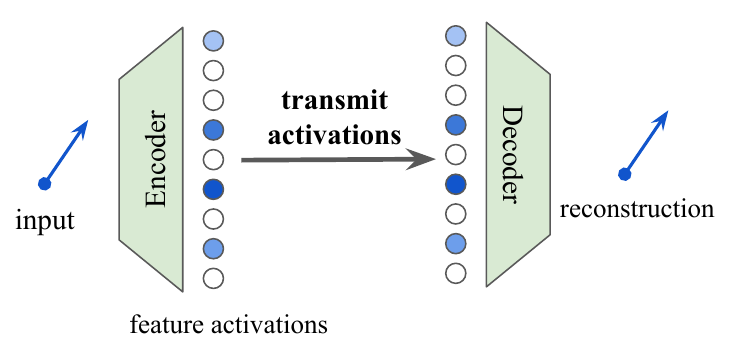}
\caption{A schematic showing a sparse autoencoder (SAE) being used to communicate an input by transmitting the encoded activations and decoding them into a reconstruction of the input.}
\end{figure}

\begin{definition}
An explanation $e$ of some phenomena $p$ is a statement $e(p)$ for which knowing $e(p)$ gives some information about $p$. An explanation is typically a natural language statement\footnotemark{}.
\end{definition}

\footnotetext{We will treat SAE activations and feature vectors as explanations themselves. Technically, we would want to do the additional step of interpreting their activation patterns or the results of causal interventions to get a natural language statement.}

\begin{definition}
The Description Length ($DL$) of some explanation $e$ is given as $|e|$, where $|\cdot|$ is the metric denoting the number of bits needed to send the explanation through a communication channel.
\end{definition}

The description length ($DL$) of an explanation is the number of bits needed to transmit the explanation. For an SAE, this would be $DL = |z|_\text{bits} + |Dec(\cdot)|_\text{bits}$. The first term is O(n) and the second term is O(1) in the dataset size so the first term dominates in the large data regime. 

\textbf{Occam's Razor}: All else equal, an explanation $e_1$ is preferred to explanation $e_2$ if $DL(e_1) < DL(e_2)$. Intuitively, the simpler explanation is the better one. We can operationalise this as the Minimal Description Length (MDL) Principle for model selection: Choose the model with the shortest description length which solves the task. It has been observed that lower description length models often generalise better \citep{mackay2003information}.

\begin{definition}
We define the Minimal Description Length (MDL) as
$MDL_\varepsilon(x) = \min  DL(SAE)$ where $Loss(x, \widehat{x}) < \varepsilon$ and $\widehat{x} = SAE(x)$.
We say an SAE is $\varepsilon$-MDL-optimal if it obtains this minimum.
\end{definition}

% \textbf{Definition 3 - Minimal Description Length (MDL)}:
% $MDL_\varepsilon(x) = \min  DL(SAE)$ where $Loss(x, \widehat{x}) < \varepsilon$ and $\widehat{x} = SAE(x)$.
% We say an SAE is $\varepsilon$-MDL-optimal if it obtains this minimum.

\section{Interpretability requires independent additivity}
Following Occam's razor we prefer simpler explanations, as measured by description length. But SAEs are not intended to simply give compressed explanations. They are also intended to give explanations that are interpretable and ideally human-understandable. 
% So we must account for how humans would actually make sense of feature activations.

SAE features can be interpreted either as \textbf{causal results} of the model inputs (which we can see by analyzing feature activation patterns) or they can be interpreted as \textbf{causes} of the model outputs (which we can see through conducting interventions on the features and seeing the downstream effects). In both cases, we want to be able to understand each SAE feature independently, without needing to control for the activations of the other features.  
If all the feature activations are causally entangled---as is the case for the dense neural activations themselves---then they are not interpretable. 
Note that for $D$ features there are $O(D^2)$ pairs of features and $\sum_i^K {D\choose{i}}$ possible sets of features which is much too large for humans to hold in working memory. So for feature explanations to be human-understandable we cannot have the all the features being entangled such that understanding a single concept requires understanding arbitrary feature interactions.

% For human interpretability, there's a good reason for this: given $D$ features there are $O(D^2)$ pairs of features and $\sum_i^K {D\choose{i}}$ possible sets of features. Humans can only typically hold a few concepts in working memory, so if features are all entangled such that understanding a single concept requires understanding arbitrary feature interactions, the explanation will not be human-understandable. This is also why the dense neural activations themselves are typically not interpretable.

Hence, for interpretability, we need to be able to understand features independently of each other such that understanding a collection of features together is equivalent to understanding all the features separately. We call this property \textbf{independent additivity}, defined below.

\begin{definition}
Independent Additivity: An explanation $e$ based on a vector of feature activations $\vec{z} =  \sum_i \vec{z_i}$ is independently additive if $e(\vec{z}) \approx \sum_i e(\vec{z_i})$.
We say that a set of features $z_i$ are independently additive if they can be understood independently of each other and the explanation of the sum of the features is the sum of the explanations of the features\footnotemark{}.
\end{definition}

% \textbf{Definition 4 - Independent Additivity}: An explanation $e$ based on a vector of feature activations $\vec{z} =  \sum_i \vec{z_i}$ is independently additive if $e(\vec{z}) \approx \sum_i e(\vec{z_i})$.
% We say that a set of features $z_i$ are independently additive if they can be understood independently of each other and the explanation of the sum of the features is the sum of the explanations of the features.

\footnotetext{Note that here the notion of summation depends on the explanation space. For natural language explanations, summation of adjectives is typically concatenation ("big" + "blue" + "bouncy" + "ball" = "The big blue bouncy ball"). For neural activations, summation is regular vector addition ($\widehat{x} = \text{Dec}(\vec{z}) = \sum_i \text{Dec}(z_i)$ ).}

% Note that here the notion of summation depends on the explanation space. For natural language explanations, summation of adjectives is typically concatenation ("big" + "blue" + "bouncy" + "ball" = "The big blue bouncy ball"). For neural activations, summation is regular vector addition ($\widehat{x} = \text{Dec}(\vec{z}) = \sum_i \text{Dec}(z_i)$ ).

\begin{figure}[t]
\centering
\includegraphics[width=0.9\textwidth]{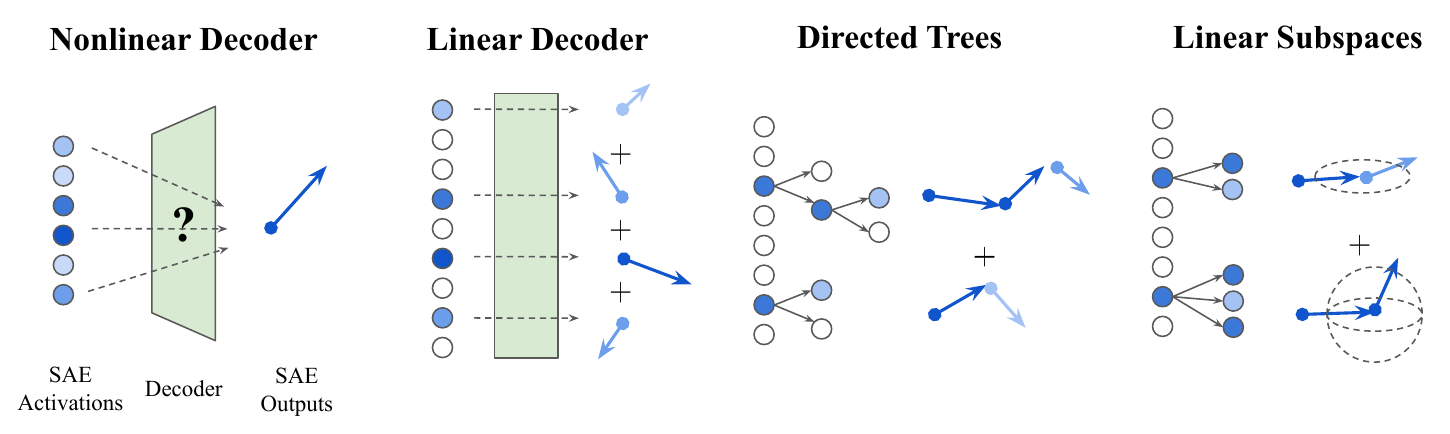}
\caption{Examples of different SAE architectures. All but nonlinear decoders are compatible with independent additivity as feature activations correspond to adding a separate vector to the output. Architectures with directed tree decoders or which allow for vectors lying within a subspace are potentially more communication efficient since a child node can only be active if its parent node is active.}
\label{fig:architectures}
\end{figure}

We see that if our SAE features are independently additive, we can also use this property for interventions and counterfactuals too. For example, if we intervene on a single feature (e.g. using it as a steering vector), we can understand the effect of this intervention without needing to understand the other features.

The independent additivity condition is directly analogous to the "composition as addition" property of the Linear Representation Hypothesis (LRH) discussed in \citet{Olah2024linear}. \textit{Independent additivity} relates to the SAE features being composable via addition with respect to the explanation - this is a property of the SAE Decoder.
In the Linear Representation Hypothesis however, \textit{Composition as Addition} is about the underlying true features (i.e. the generating factors of the underlying distribution), which is a property of the underlying distribution.

% \begin{figure}[t]
% \centering
% \includegraphics[width=0.6\textwidth]{assets/hierarchical_independent_additivity.png}
% \caption{Schematic of SAEs as a communication protocol}
% \end{figure}

It is immediate from the definition that Independent Additivity holds for linear decoders however, we note that this condition also allows for more general decoder architectures. For example, features can be arranged to form a collection of directed trees, shown in \cref{fig:architectures}, where arrows represent the property "the child node can only be active if the parent node is active"\footnotemark{}. Here each feature still corresponds to its own vector direction in the decoder. Since each child feature has a single path to its root feature, there are no interactions to disentangle and the independent additivity property still holds, in that each \textit{tree} can be understood independently in a way that's natural for humans to understand, as a multi-dimensional feature. An advantage of the directed-tree SAE decoder structure is that it can be more description-length efficient as shown in \cref{fig:hierarchical}.

\footnotetext{In practice, we typically expect feature trees to be shallow structures which capture causal relationships between highly related features. A particularly interesting example of this structure is a group-sparse autoencoder where linear subspaces are densely activated together.}

% In practice, we typically expect feature trees to be shallow structures which capture causal relationships between highly related features. A particularly interesting example of this structure is a group-sparse autoencoder where linear subspaces are densely activated together.

Independent additivity of feature explanations also implies that the description length of the set of activations, $\{z_i\}$, is the sum of the lengths for each feature $DL(\{z_i\}) = \sum_i DL(z_i)$. If we know the distribution of the activations, $p_i(z)$, then it is possible to send the activations using an average description length equal to the distribution's entropy, $DL(z_i) = H(p_i) \equiv \sum_{z\in Z} -p_i(z) \log_2 p_i(z)$.  For directed trees, the average description length of a child feature would be the conditional entropy, $DL_\text{child}(z_i) = H(p_i | \text{ parent active})$, which accounts for the fact that $DL = 0$ when the parent is not active. This is one reason that directed tree-style SAEs can potentially have smaller descriptions than conventional SAEs.

% Powerful nonlinear autoencoders could potentially compress activations further and reduce the minimum description length but are not consistent with independent additivity. The compressed activations would likely be uninterpretable since there are interactions between the $z_i$ that remain entangled. It may not be generally possible to read off the effects of a single feature activation by looking only at that feature. For an arbitrary non-linear decoder, interpreting $z_i$ depends on all $z_{j\neq i}$\footnotemark{}.

% \footnotetext{One possible middle ground here might be bilinear structures which can be more expressive than linearity but have been shown to maintain some interpretability properties as in \citet{sharkey2023technical}.}

\section{SAEs should be sparse, but not too sparse}

Naively we might see SAEs as decompressing neural activations which contain densely packed features in superposition. To see that SAEs are producing compressed explanations of activations we must note that the inherent feature sparsity means that it is more efficient to communicate SAE latent features rather than neural activations even though the dimension of the latent dimension is higher.

The description length for a set of SAE activations (under independent additivity) with distribution $p(z)$ is given by $H(p) = \sum_{z\in Z} -p(z) \log_2 p(z)$. For exposition, consider a simpler formulation where we directly consider the bits needed without prior knowledge of the distributions.  For a set of feature activations with $L_0$ nonzero elements out of $D$ dictionary features, an upper bound on the description length is \begin{align}
    DL \lesssim L_0 (B + \log_2 D)
\end{align} where $B$ is the effective precision of each float and $\log_2D$ is the number of bits required to specify which features are active. To achieve the same loss, higher sparsity (lower $L_0$) typically requires a larger dictionary, so there's an inherent trade-off between decreasing L0 and decreasing the dictionary size in order to reduce description length. 

As an illustrative example, we compare reasonable hyperparameters for GPT-2 SAEs to dense/narrow and sparse/wide extreme hyperparameters:

% We show that an SAE \citep{bloom2024open} has a description length of ~1,405 bits per input token, compared to 5,376 bits for transmitting the dense neural activations and 13,993 bits for a one-hot encoding of all possible token sequences of length 128. Here the SAE at intermediate sparsity and width has the lower description length. 

\begin{itemize}
\item \textbf{Reasonable SAEs}: \citet{bloom2024open}'s open-source SAEs for GPT-2 layer 8 have $L_0=65$, $D=25,000$. Given $B = 7$ bits per nonzero float (8-bit quantization with the sign fixed to positive), the description length per input token is 1405 bits.

\item \textbf{Dense Activations}: A dense representation that still satisfies independent additivity would be to send the neural activations directly instead of training an SAE. GPT-2 has a model size of $d=768$, the description length is simply $DL$ = B d = 5376 bits per token.

\item \textbf{One-hot encodings}: At the sparse extreme, our dictionary has a row for each neural activation in the dataset, so $L_0 = 1$ and $D = (\text{vocab size})^{\text{seq len}}$. GPT-2 has a vocab size of 50,257 and the SAEs are trained 128 token sequences. All together this gives $DL = 13,993$ bits per token.
\end{itemize}

Although the comparison is slightly unfair because the SAE is lossy (93\% variance explained) and the other cases are lossless, these calculations demonstrate that reasonable SAEs are indeed compressed compared to the dense and sparse extremes.
We hypothesise that the reason that we're able to get this helpful compression is that the true features from the generating process are themselves sparse. 

Note the difference here from choosing models based on the reconstruction loss vs sparsity ($L_0$) Pareto frontier. When minimising $L_0$, we are encouraging decreasing $L_0$ and increasing $D$ until $L_0=1$. Under the MDL model selection paradigm we are typically able to discount trivial solutions like a one-hot encoding of the input activations and other extremely sparse solutions which make the reconstruction algorithm analogous to a k-Nearest Neighbour.
% classifier\footnotemark{}.

% \footnotetext{Note that we cannot always strictly rule out these solutions since there is some dependency on the loss tolerance $\varepsilon$ given and the dataset. We show how this plays out for a real dataset in the following section.}

\section{MDL-SAEs find interpretable and composable features for MNIST}

\citet{lee2001introduction} describe the classical method for using the Minimal Description Length (MDL) criteria for model selection. Here we choose between model hyperparameters (in particular the SAE width and expected $L_0$) for the optimal SAE. Our algorithm for finding the MDL-SAE solution and details for this case study are given in Appendix C.

\begin{figure}[t]
\centering
\includegraphics[width=1.0\textwidth]{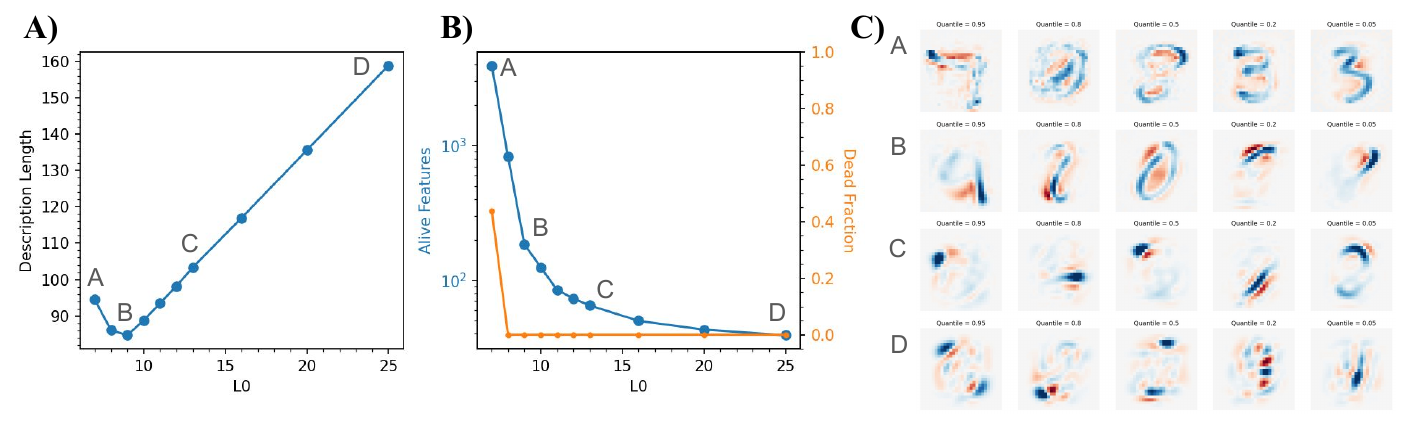}
\caption{Finding the minimal description length (MDL) solution for SAEs trained on MNIST. A) Description length vs sparsity ($L_0$) for a set of hyperparameters with the same reconstruction error.  B) Plot of the number of alive features as a function of sparsity ($L_0$). C) A random sample of SAE features at the 95th, 80th, 50th, 20th, and 5th percentiles of feature density respectively.}
\label{fig:mnist_graphic}
\end{figure}

We trained SAEs on the MNIST dataset of handwritten digits \citep{lecun1998gradient} and find the set of hyperparameters resulting in the same test MSE. We see three basic regimes:
\begin{itemize}
    \item \textbf{High $L_0$, narrow SAE width} (C, D in \cref{fig:mnist_graphic}): Here, the description length (DL) is linear with $L_0$, suggesting that the DL is dominated by the number of bits needed to represent the $L_0$ nonzero floats. The features appear as small sections of digits that could be relevant to many digits (C) or start to look like dense features that one might obtain from PCA (D).
    \item \textbf{Low $L_0$, wide SAE width} (A in \cref{fig:mnist_graphic}): Though $L_0$ is small, the DL is large because as the SAE becomes wider, additional bits are required to specify which activations are nonzero. The features appear closer to being full digits, i.e. similar to samples from the dataset. 
    % Note that the features appear somewhat noisy because early stopping was needed to prevent overfitting to the train set.
    \item \textbf{The MDL solution} (B in \cref{fig:mnist_graphic}): There's a balance between the two contributions to the description length. The features appear like longer line segments or strokes for digits, but could apply to multiple digits. 
\end{itemize}
In this example, the MDL solution finds a meaningful decomposition of digits into stroke-like features. 
More dense SAEs find less interpretable point-like features, while sparser SAEs find features that resemble examples from the dataset and fail to decompose the digits into reusable and composable features. 

\section{Optimising for MDL can reduce undesirable feature splitting}

In large language models, SAEs with larger dictionaries learn finer-grained versions of features learned in smaller SAEs, a phenomenon known as "feature splitting" \citep{bricken2023monosemanticity}. Feature splitting that introduces a novel conceptual distinction is desirable but some feature splitting---for example, learning dozens of features representing the letter "P" in different contexts \citep{bricken2023monosemanticity}---is undesirable and can waste dictionary capacity while not giving more explanatory power. 

A toy model of undesirable feature splitting is an SAE that represents the AND of two boolean features, $A$ and $B$, as a third feature direction. The two booleans represent whether the feature vectors $v_A$ and $v_B$ are present or not, so there are four possible activations: $0$, $v_A$, $v_B$, and $v_A+v_B$.

\begin{figure}[t]
\centering
\includegraphics[width=0.9\textwidth]{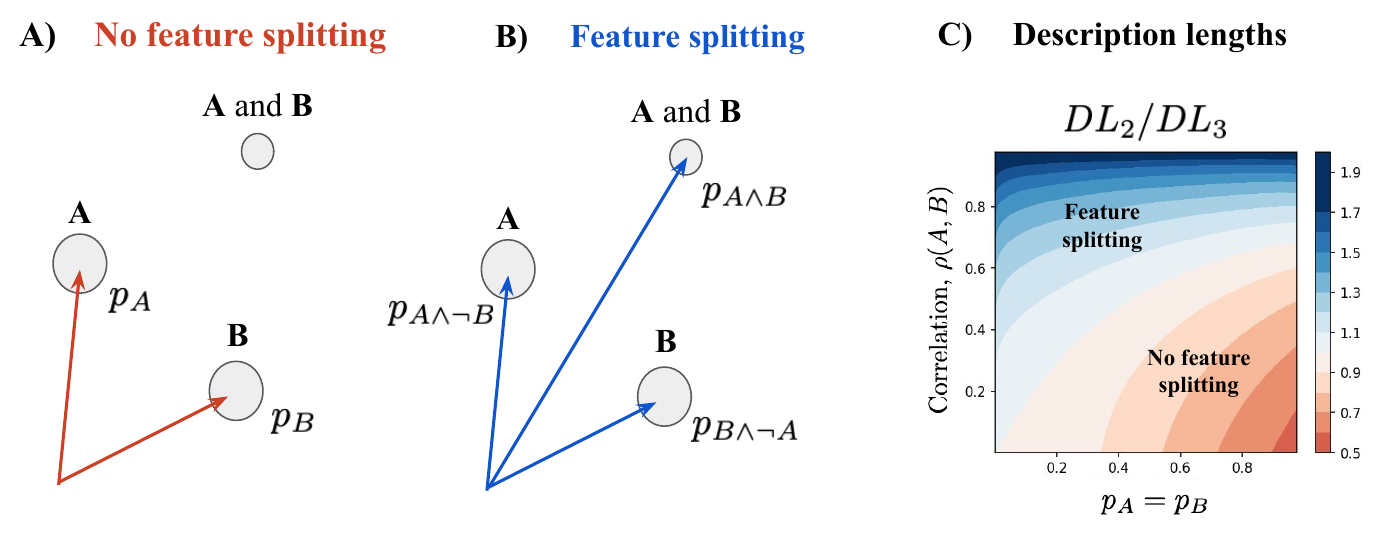}
\caption{A toy model of undesirable feature splitting. The SAE can learn two boolean features without feature splitting (A) or three mutually exclusive boolean features with feature splitting (B) which always has lower $L_0$. Minimizing description length provides a decision boundary (C) for when feature splitting is preferred or not. \vspace{-0.5cm}}
\label{fig:feature_splitting}
\end{figure}
\textbf{No Feature Splitting}: Say that the SAE only learns two boolean feature vectors, $v_A$ and $v_B$, as shown in \cref{fig:feature_splitting}. It is still capable of reconstructing $A\wedge B$ as the sum $v_A + v_B$. The $L_0$ would simply be the expectation of the boolean activations, so $L_0 = p_A + p_B$ and the description length would be $DL = H(p_A) + H(p_B)$ where $H(p)$ is the entropy of a Bernoulli variable with probability $p$.

\textbf{Feature Splitting}: In this case, the SAE learns three mutually exclusive features. $A\wedge B$ is explicitly represented with the vector $v_A + v_B$ while the two other features represent $A\wedge \neg B$ and $B \wedge \neg A$ with vectors $v_A$ and $v_B$. This setup has the same reconstruction error but has lower $L_0 = p_{A\wedge\neg B} + p_{B\wedge\neg A} + p_{A\wedge B} = p_A + p_B - p_{A\wedge B}$ since the probabilities for $A\wedge \neg B$, say, are reduced as $p_{A\wedge \neg B} = p_A - p_{A\wedge B}$. Note that the $L_0$ (sparsity) is necessarily lower than in the non-feature splitting case.
 % The description length, however is now given as $DL = H(p_A) + H(p_B) + H(p_{A\wedge B})$ by independent additivity.

% \footnotetext{With three features in a 2d plane it may seem difficult to find a linear encoder to separate them, but we can find lines that separate each feature from the other two and have the encoder measure the distance away from dividing line.}

Even though feature splitting always results in a lower $L_0$, it does not always result in the smallest description length. The phase diagram in \cref{fig:feature_splitting} shows the case where $p_A = p_B$. If the correlation coefficient $\rho$ between $A$ and $B$ is small then representing only $A$ and $B$, but not $A\wedge B$, takes fewer bits so the preferred solution avoids feature splitting. However, if the correlation is large, then feature splitting is preferred since $A\wedge B$ occurs frequently enough that explicitly representing it reduces the description length. In this way, minimizing description length can limit the amount of undesirable feature splitting and gives us a concrete decision criteria to understand when we might expect feature splitting.

% Imagine that we have an SAE with fixed width and small $\varepsilon$ loss but one of the features is as of yet undecided. We might choose between representing some scarcely used direction which explains some variance or the AND of two features that are already in our feature codebook (an example of feature splitting). This is a problem of deciding which will give the largest improvement on the loss-$DL$ Pareto curve. Adding the novel feature will likely improve the loss but adding the composite feature may improve the description length. This tradeoff will lead to a stricter condition on when to add the composite feature than suggested in the phase diagram above.

\section{Hierarchical features allow for more efficient coding schemes}

\begin{figure}[h]
\centering
\includegraphics[width=1.0\textwidth]{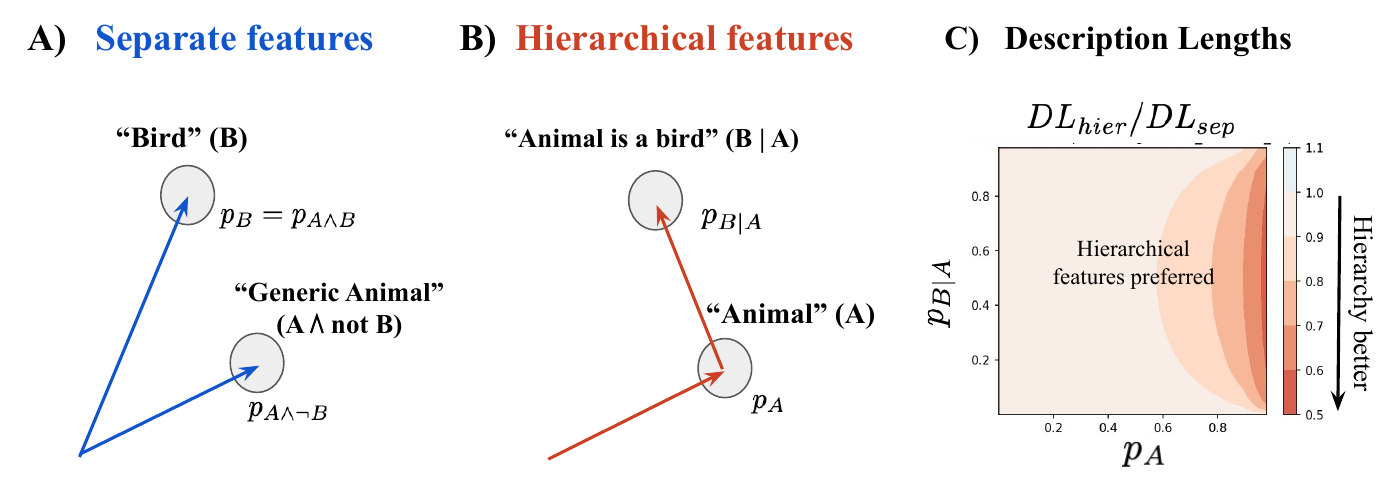}
\caption{Two naturally hierarchical boolean features, such as "Animal" and "Bird", can be learned as separate mutually exclusive features (A) or in hierarchy (B) where the child feature can only be active if the parent feature is active, captured by the conditional probability $p_{B|A}$. C) The hierarchical case always has lower description length (DL) since the child feature's activations need not be sent when the parent is not active.}
\label{fig:hierarchical}
\end{figure}

Often features are semantically or causally related and this should allow for more efficient coding schemes. 
For example, consider the hierarchical concepts "Animal" ($A$) and "Bird" ($B$). Since all birds are animals, the "Animal" feature will always be active when the "Bird" feature is active. A conventional SAE would represent these as separate feature vectors, one for "Bird" ($B$) and one for "Generic Animal" ($A\wedge \neg B$), that are never active together, as shown in \cref{fig:hierarchical}. This setup has a low $L_0$, equal to the probability of "Animal", $p_A$, since something is a bird, a generic animal, or neither.

An alternative approach would be to define a variable length coding scheme \citep{salomon2007variable}. For example, one might consider first sending the activation for "Animal" ($A$) and only if "Animal" is active, sending the activation for "Animal is a Bird" ($B|A$). Now the description length is given as $DL = H(p_A) + p_A H(p_{B|A})$ which is always fewer bits compared to the conventional SAE with $DL = H(p_A - p_B) + H(p_B)$, (see the phase diagram in \cref{fig:hierarchical}). The overall $L_0$ however is higher because sometimes two activations are nonzero at the same time, so $L_0 = p_A + p_{B|A}$. 

This case illustrates the potential to reduce description length by matching the SAE architecture more closely to the hierarchical and causal structure of the data distribution. We also see another case where optimising for sparsity differs to the MDL approach - hierarchical structures of the type described above are never beneficial when optimising for sparsity but when thinking in terms of Description Length, there are clear benefits to using the semantic structure of the data. 

% In order to make use of this coding scheme, the encoder and decoder would need to explicitly have the same hierarchical structure. Otherwise, it would not be easy to identify the dependencies between features when assessing the description length.

% Note: Hierarchical features are a special case of having ANDs of features, discussed in the Feature Splitting section above. For example, we could define "Bird" in terms of its specific properties as "Bird" equals "Animal" AND "Has Wings" AND "Has Beak" etc. These properties are highly correlated with "Animal" and each other, which is why it's possible to define a more efficient coding scheme.

\section{Related Work}

\citet{bricken2023features} also consider how information measures relate to SAEs and find that "bounces" in entropy correspond to dictionary sizes with the correct number of features in synthetic experiments. We find a similar bounce in description length in a non-synthetic experiment. We go further by studying several examples where minimal description length gives more intuitive features and discuss more description-efficient SAE architectures. 
% discuss defining features as "the simplest factorization of the activations". We don't directly claim that this should be the definition of features but we instead argue that features make up explanations, where simpler explanations are preferred. Given our experiments we find similar "bounces" (global minima) in the description length as a function of SAE width, rather than a monotonic function. Larger dictionaries tend to require more information to represent, but sparser codes require less information to represent, which may counterbalance. 

Our setting is inspired by rate-distortion theory \citep{shannon1948mathematical} and the Rate-Distortion-Perception Tradeoff \citep{blau2019rethinking}, which notes the surprising result that distortion (e.g. squared-error distortion) is often at odds with perceptual quality and suggest that the divergence $d(p_X, p_{\hat X})$ more accurately represents perception as judged by humans (though the exact divergence which most closely matches human intuition is still an ongoing area of research).

As in \citet{ramirez2012mdl}, we use the MDL approach for the Model Selection Problem using the criteria that the best model for the data is the model that captures the most useful structure from the data. 
The more structure or "regularity" a model captures, the shorter the description of the data, X, will be under that model (by avoiding redundancy in the description). Therefore, MDL will select the best model as the one that
produces the most efficient description of the data.
\citet{chan2024compact} use Mechanistic Interpretability techniques to generate compact formal guarantees (i.e. proofs) of model performance and also note a deep connection between interpretability and compression.

\citet{dhillon2011minimum} use the information theoretic MDL principle to motivate their Multiple Inclusion Criterion (MIC) for learning sparse models. Their setup is similar to ours but their method relies on sequential greedy-sampling rather than a parallel approach, which performs slower than the SAE methods on modern hardware but is otherwise a promising approach. They present applications where human interpretability is a key driver of the reason for a sparse solution and we present additional motivations for sparsity as plausibly aligning with human interpretability.

\citet{sharkey_sparsify_2024} motivates a high-level framework for Mechanistic Interpretability involving three stages: mathematical description (breaking the network down into functional parts), semantic description (labelling the functional parts) and validation (using the semantic description to make predictions about model behaviour and evaluating these predictions). Here we focus on the mathematical description stage, trading off mathematical description length with mathematical accuracy (or faithfulness) to the network's representations.

\citet{chan2024compact} use Mechanistic Interpretability techniques to generate compact formal guarantees (i.e. proofs) of model performance. Here they are seeking \textit{explanations} which bound the model loss by some $\varepsilon$ on a task. They find that better understanding of the model leads to shorter (i.e. lower description length) proofs. Similar to our work the authors note the deep connection between mechanistic interpretability and compression.

\section{Conclusion}

In this work, we have presented an information-theoretic perspective on Sparse Autoencoders as explainers for neural network activations. Using the MDL principle, we provide some theoretical motivation for existing SAE architectures and hyperparameters. We also hypothesise a mechanism for, and criteria to describe, the commonly observed phenomena of feature splitting. In the cases where feature splitting can be seen as undesirable for downstream applications, we hope that, using this theoretical framework, the prevalence of undesirable feature splitting could be decreased in practical modelling settings.

A limitation of this work as presented is that the MDL priniciple is treated as strategy for model selection: to choose a model out of a collection of multiple models. However, training multiple models with a hyperparameter sweep may be computationally expensive. Future work could look to include the entropy term in the loss function and optimise for it directly through either a straight-thought estimation approach or with a Bayesian prior. 

Our work suggests a path to a formal link between existing interpretability methods and information-theoretic principles such as the Rate-Distortion-Perception trade-off and two-part MDL coding schemes. We would be excited about work which further connects concise explanations of learned representations to well-explored problems in compressed sensing.

Historically, evaluating SAEs for interpretability has been difficult without human interpretability ratings studies, which can be labour intensive and expensive. We propose that operationalising interpretability via description length can help in creating principled evaluations for interpretability, requiring less subjective and expensive SAE metrics. 
% We expect minimizing description length to align with human-interpretability since it yields concise explanations using independently understandable features. The features may be model- and task-specific and not correspond to existing human concepts but are likely still human-learnable.

We would be excited about future work which explores to what extent variants in SAE architectures can decrease the MDL of communicated latent feature activations. In particular, we suggest that exploiting causal structure inherent in the data distribution may be important to efficient coding. 
% We would also be interested in future work which explores the relationship between the MDL-optimal hyperparameters for a given allowable error rate $\varepsilon$, possibly through scaling laws analysis.

\vfill

\pagebreak 

% \section*{References}

\bibliography{references.bib}
\bibliographystyle{abbrvnat}

\vfill

%%%%%%%%%%%%%%%%%%%%%%%%%%%%%%%%%%%%%%%%%%%%%%%%%%%%%%%%%%%%
\pagebreak

\appendix

\section{Details on determining the MDL-SAE}

\subsection{Algorithm}
\begin{enumerate}
    \item \textbf{Specify a tolerance level, $\varepsilon$, for the loss function}. The tolerance $\varepsilon$ is the maximum allowed value for the loss, either the reconstruction loss (MSE for the SAE) or the model's cross-entropy loss when intervening on the model to swap in the SAE reconstructions in place of the clean activations. For small datasets using a reconstruction, the test loss should be used. 
    % <!-- We suggest using cross-entropy loss. -->
    \item \textbf{Train a set of SAEs within the loss tolerance}. It may be possible to simplify this task by allowing the sparsity parameter to also be learned.
    \item \textbf{Find the effective precision needed for floats}. The description length depends on the float quantisation. We typically reduce the float precision until the change in loss results in the reconstruction tolerance level is exceeded.
    \item \textbf{Calculate description lengths}. With the quantised latent activations, the entropy can be computed from the (discretized) probability distribution, $\{p^i_\alpha\}$, for each feature $i$, as $$H = \sum_{i,\alpha} -p^i_\alpha \log p^i_\alpha$$
    \item \textbf{Select the SAE that minimizes the description length} i.e. the $\varepsilon$-MDL-optimal SAE.
\end{enumerate}

\subsection{Details for MNIST case study}
For MNIST, we trained BatchTopK SAEs \citep{bussmann2024batch}, typically for 1000+ epochs until the test reconstruction loss converged or stopping early in cases of overfitting.  Our desired MSE tolerance was $0.0150$. Discretizing the floats to roughly 5 bits per nonzero float gave an average change in MSE of $\approx 0.0001$, which was roughly the scale over which MSE varied for the hyperparameters used. 

\citet{gao2024scaling} find that as the SAE width increases, there's a point where the number of dead features starts to rise. In our experiments, we noticed that this point seems to be at a similar point to where the description length starts to increase as well, although we did not test this systematically and this property may be somewhat dataset dependent.

%%%%%%%%%%%%%%%%%%%%%%%%%%%%%%%%%%%%%%%%%%%%%%%%%%%%%%%%%%%%

\end{document}